\documentclass[10pt]{article} 

\usepackage[preprint]{tmlr}

\usepackage{amsmath,amsfonts,bm}

\def\eqref#1{equation~\ref{#1}}

\def\1{\bm{1}}

\DeclareMathAlphabet{\mathsfit}{\encodingdefault}{\sfdefault}{m}{sl}
\SetMathAlphabet{\mathsfit}{bold}{\encodingdefault}{\sfdefault}{bx}{n}

\usepackage{hyperref}
\usepackage{url}
\usepackage{graphicx}
\usepackage{booktabs}
\usepackage{float}
\usepackage{enumitem}

\setlist[itemize]{noitemsep, topsep=2pt, parsep=0pt, partopsep=0pt}
\title{When Consistency Does Not Mean Reliability:
Evaluating Local LLM Judges Against Human Ratings}

\author{Aakash Kumar Tiwari \\
Department of Mathematics \\
Indian Institute of Technology Kharagpur \\
\texttt{tiwariaakash1025@kgpian.iitkgp.ac.in}}

\def\month{MM}  
\def\year{YYYY} 
\def\openreview{\url{https://openreview.net/forum?id=XXXX}}

\begin{document}

\maketitle


\footnotetext{AI writing assistance was used during manuscript
preparation for language editing, organization, and clarity. The study
design, experiments, analysis, results, and scientific claims were
developed and verified by the authors.}


\begin{abstract}
Large language models (LLMs) are increasingly used to evaluate the responses of other language models. This approach, known as LLM-as-a-Judge, is faster and cheaper than human evaluation. However, a judge may produce consistent scores without necessarily agreeing with human evaluators. In this work, we study this issue using two local open-weight LLM judges, LLaMA-3-8B and Qwen2.5-7B. We evaluate 300 responses generated by an instruction-tuned GPT-2 (124M) model for 100 questions covering five categories: factual knowledge, instruction following, mathematics, reasoning, and writing. Each response is scored by nine human annotators and is evaluated three times by each LLM judge using the same rubric. We compare the judge
scores with the average human scores using Pearson correlation, Spearman correlation, mean absolute error (MAE), signed bias, and self-consistency. LLaMA-3-8B shows a Pearson correlation of 0.275 with human scores, while Qwen2.5-7B achieves 0.340. Their MAEs are 27.71 and 18.64, respectively. Despite this limited agreement,
both judges show high self-consistency, with exact consistency rates of 97.3\% for LLaMA-3-8B and 92.3\% for Qwen2.5-7B. These results show that high self-consistency does not necessarily indicate high agreement with human judgments. Our findings highlight the need to evaluate both consistency and human alignment when using local LLMs as automatic judges.
\end{abstract}


\section{Introduction}
Large language models (LLMs) have achieved strong performance across tasks such as question answering, summarization, reasoning, and instruction following. As their capabilities increase, evaluating generated responses has become an important research problem. Traditional metrics such as BLEU, ROUGE, and BERTScore are useful for
text generation but may not adequately capture correctness, instruction following, reasoning quality, and overall response quality \cite{papineni2002bleu, lin2004rouge, zhang2020bertscore}. Large-scale benchmarks such as BIG-bench and PromptBench further emphasize the need for systematic evaluation of modern language models \cite{srivastava2023beyond,zhu2024promptbench}.
Human evaluation remains an important reference for open-ended responses because human evaluators can assess correctness, relevance, clarity, and instruction adherence. However, human evaluation is expensive, time-consuming, and difficult to scale. Platforms such as Chatbot Arena demonstrate the value of human judgments at scale, but large-scale annotation still requires substantial effort
\cite{chiang2024chatbotarena}. These limitations have motivated the use
of language models as automatic evaluators.
The use of an LLM to evaluate another model is commonly referred to as
\emph{LLM-as-a-Judge}. Zheng et al.\ showed that capable LLMs can serve
as judges for open-ended evaluation and achieve substantial agreement
with human preferences \cite{zheng2023judging}. G-Eval similarly demonstrated that GPT-4-based evaluation can achieve strong human alignment on several natural language generation tasks \cite{liu2023geval}. These results establish LLM-based judging as a scalable alternative to fully manual evaluation.
However, using an LLM as a judge introduces a separate question: \emph{how reliable is the judge itself?} A judge may assign similar scores when the same response is evaluated repeatedly. We refer to this property as \emph{self-consistency}. Such internal stability, however, does not necessarily imply agreement with independent human evaluators. A judge may therefore be highly stable while systematically
assigning scores that differ from human judgments. This issue is particularly relevant for local and open-weight judges. JudgeLM investigated fine-tuned language models as scalable judges, while Prometheus and Prometheus 2 developed open evaluators for fine-grained and rubric-based assessment
\cite{zhu2023judgelm,kim2023prometheus,kim2024prometheus2}. FLASK further
explored fine-grained evaluation through alignment skill sets
\cite{ye2024flask}. These studies demonstrate the potential of open-weight evaluators while also indicating that judge behavior can depend on the model, evaluation criterion, and task.
Recent studies have raised further concerns about human--judge agreement and evaluation bias. Huang et al.\ found that fine-tuned judges may perform well in particular settings without being general substitutes for stronger judges such as GPT-4 \cite{huang2024empirical}. Chen et al.\ showed that both humans and
LLM judges can exhibit systematic judgment biases \cite{chen2024humans}. JudgeBench emphasized the importance of evaluating the judges themselves \cite{tan2025judgebench}, while Bavaresco et al.\ reported substantial variation in agreement between LLM judges and humans across models and NLP tasks
\cite{bavaresco2025llms}.
Despite this progress, an important question remains: \emph{does a highly consistent LLM judge necessarily provide reliable judgments from a human perspective?} Existing work has examined human alignment, judge biases, open and fine-tuned evaluators, and judge benchmarking. Our study focuses specifically on separating
\emph{self-consistency} from \emph{human alignment}. Repeated agreement between a judge's own scores measures the stability of its evaluation process, whereas comparison with independent human ratings measures its alignment with human judgments.
To investigate this distinction, we construct a controlled evaluation
setting using an instruction-tuned GPT-2 (124M) model. We evaluate 300 generated responses from 100 questions across five categories: factual knowledge, instruction following, mathematics, reasoning, and writing. Each response is independently scored by nine human annotators and evaluated three times by two local open-weight judges,
LLaMA-3-8B and Qwen2.5-7B, using the same explicit scoring rubric. This design allows us to measure both human--judge agreement and repeated judge consistency.
The results show a clear difference between these properties. LLaMA-3-8B achieves a Pearson correlation of $0.275$ with human scores, while Qwen2.5-7B achieves $0.340$. In contrast, both judges show high repeatability, with exact consistency rates of $97.3\%$ and $92.3\%$, respectively. Thus, a judge can remain highly stable in repeated evaluations while showing limited agreement with human evaluators.
The main contributions of this work are as follows:

\begin{itemize}
    \item We provide a controlled empirical study of two local
    open-weight LLM judges, LLaMA-3-8B and Qwen2.5-7B, using a common
    rubric and repeated evaluation protocol.
    \item We explicitly distinguish \emph{self-consistency} from
    \emph{human alignment} using repeated judge evaluations and
    independent human ratings.
    \item We evaluate human--judge agreement using Pearson correlation,
    Spearman correlation, MAE, signed bias, and self-consistency.
    \item We analyze judge behavior across question categories and
    response-generation decoding configurations.
    \item We examine response length as a potential factor by comparing
    raw and length-controlled human--judge correlations.
\end{itemize}
Overall, this study highlights that \emph{consistency alone should not be treated as evidence of reliability}. Local LLM judges should be evaluated using both repeated-evaluation consistency and agreement with independent human judgments.

\section{Related Work}
The evaluation of language model outputs has traditionally relied on automatic metrics and human judgments. With increasingly capable language models, LLM-based evaluation has become an important research direction. This section reviews automatic evaluation metrics, LLM-as-a-Judge methods, open and fine-tuned evaluators, and human alignment and judge reliability.

\subsection{Automatic Evaluation of Generated Text}
Early evaluation of generated text mainly relied on reference-based metrics. BLEU measures modified n-gram precision \cite{papineni2002bleu}, while ROUGE evaluates lexical overlap for generated summaries \cite{lin2004rouge}. BERTScore later introduced contextual representations from pretrained language models to measure
semantic similarity \cite{zhang2020bertscore}.
Although computationally efficient, these metrics may not adequately capture factual correctness, instruction following, reasoning quality, or other qualitative properties. This limitation is important for open-ended generation, where valid responses may differ substantially in wording. BIG-bench and PromptBench support systematic evaluation across diverse capabilities and prompting settings
\cite{srivastava2023beyond,zhu2024promptbench}. Sottana et al.\ further showed that model-based evaluation can be useful for sequence-to-sequence tasks while exhibiting task-dependent variation \cite{sottana2023evaluation}. These limitations motivated more flexible evaluators for assessing multiple response properties.

\subsection{LLM-as-a-Judge}
LLM-as-a-Judge uses a language model to evaluate the outputs of another language model, typically with criteria or a rubric covering properties such as correctness, relevance, coherence, and instruction following. Zheng et al.\ systematically studied LLM judges using MT-Bench and Chatbot Arena and showed that capable LLMs can provide scalable evaluation while exhibiting systematic biases \cite{zheng2023judging}. G-Eval used GPT-4 with explicit criteria and forms for reference-free natural language generation evaluation, demonstrating improved human alignment \cite{liu2023geval}. However, these studies do not directly establish whether smaller local models provide similarly reliable judgments.
Human preference evaluation remains an important reference for validating automated evaluators. Chatbot Arena uses large-scale human pairwise preferences to compare language models \cite{chiang2024chatbotarena}, while AlpacaFarm provides a framework
for collecting and simulating human preference data for language model alignment \cite{dubois2023alpacafarm}.

\subsection{Open and Fine-Tuned LLM Judges}
The cost of proprietary models has motivated research on open and fine-tuned evaluators. JudgeLM investigated fine-tuned language models as scalable judges and analyzed evaluation biases \cite{zhu2023judgelm}. Prometheus introduced an open evaluator for fine-grained, rubric-based assessment with customizable criteria
\cite{kim2023prometheus}. Prometheus 2 extended this approach to direct assessment and pairwise ranking using user-defined criteria \cite{kim2024prometheus2}. FLASK proposed skill-based fine-grained evaluation based on individual alignment capabilities \cite{ye2024flask}.
These studies demonstrate the scalability and flexibility of open evaluators while indicating that performance can depend on the model, task, scoring criterion, and prompt. Therefore, stable scores from an open judge do not by themselves establish human-aligned reliability.

\subsection{Human Alignment, Bias, and Judge Reliability}
A central issue in LLM-based evaluation is the relationship between judge scores and human judgments. Huang et al.\ found that fine-tuned judges can perform well in particular settings without being general substitutes for stronger evaluators such as GPT-4 \cite{huang2024empirical}. Chen et al.\ showed that both humans and LLM
judges can exhibit systematic judgment biases and sensitivity to factors
unrelated to response quality \cite{chen2024humans}.
Recent work has increasingly focused on evaluating the judges themselves. JudgeBench introduced a benchmark for assessing LLM-based judges on challenging evaluation cases \cite{tan2025judgebench}. Bavaresco et al.\ conducted a large-scale study across 20 NLP evaluation tasks and reported substantial variation
across models, tasks, and evaluation properties \cite{bavaresco2025llms}. Leiter et al.\ additionally emphasized the importance of explainable evaluation metrics, since numerical scores alone may not reveal the basis of an evaluation
\cite{leiter2024explainable}.

\subsection{Research Gap}
Existing work has examined human alignment, judge bias, open evaluators, and judge benchmarking \cite{zheng2023judging,huang2024empirical,chen2024humans,
tan2025judgebench,bavaresco2025llms}. However, an important distinction remains between the \emph{stability} of a judge's decisions and its \emph{agreement} with independent human judgments.
Our study explicitly evaluates these two properties of local LLM judges. We repeatedly evaluate the same responses using the same rubric and compare the resulting scores with independent human ratings. This design tests whether a judge can remain highly consistent across repeated evaluations while still showing limited agreement with humans. We therefore treat self-consistency as a property that should be measured separately from human alignment rather than as an implicit
indicator of evaluator reliability.

\begin{table}[H]
\centering
\caption{Representative studies on LLM-based evaluation and
LLM-as-a-Judge.}
\label{tab:previous_judges}
\renewcommand{\arraystretch}{1.15}
\begin{tabular}{p{2.6cm} p{2.3cm} p{3.1cm} p{3.4cm}}
\hline
\textbf{Study} & \textbf{Evaluator} &
\textbf{Evaluation Setting} & \textbf{Main Focus} \\
\hline

Zheng et al. \cite{zheng2023judging} &
Strong LLM judges &
MT-Bench and Chatbot Arena &
Human preference alignment and judge biases \\

Liu et al. \cite{liu2023geval} &
GPT-4 &
NLG evaluation &
Human-aligned reference-free evaluation \\

Huang et al. \cite{huang2024empirical} &
Fine-tuned LLM judges &
Multiple evaluation dimensions &
Generalization and limitations of fine-tuned judges \\

Zhu et al. \cite{zhu2023judgelm} &
JudgeLM &
LLM response evaluation &
Scalable open-source judging and judge biases \\

Kim et al. \cite{kim2023prometheus} &
Prometheus &
Rubric-based evaluation &
Fine-grained and customizable evaluation \\

Ye et al. \cite{ye2024flask} &
FLASK &
Skill-based LLM evaluation &
Fine-grained alignment skill assessment \\

Kim et al. \cite{kim2024prometheus2} &
Prometheus 2 &
Direct and pairwise evaluation &
Open-source evaluator models \\

Chen et al. \cite{chen2024humans} &
Humans and LLM judges &
Multiple evaluation settings &
Judgment bias and robustness \\

Tan et al. \cite{tan2025judgebench} &
Multiple LLM judges &
JudgeBench &
Reliability of LLM-based judges \\

Bavaresco et al. \cite{bavaresco2025llms} &
Multiple LLMs &
20 NLP evaluation tasks &
Agreement between LLM judges and humans \\

\hline
\end{tabular}
\end{table}
\section{Methodology}
\begin{figure}[H]
    \centering
    \includegraphics[width=0.6\textwidth]{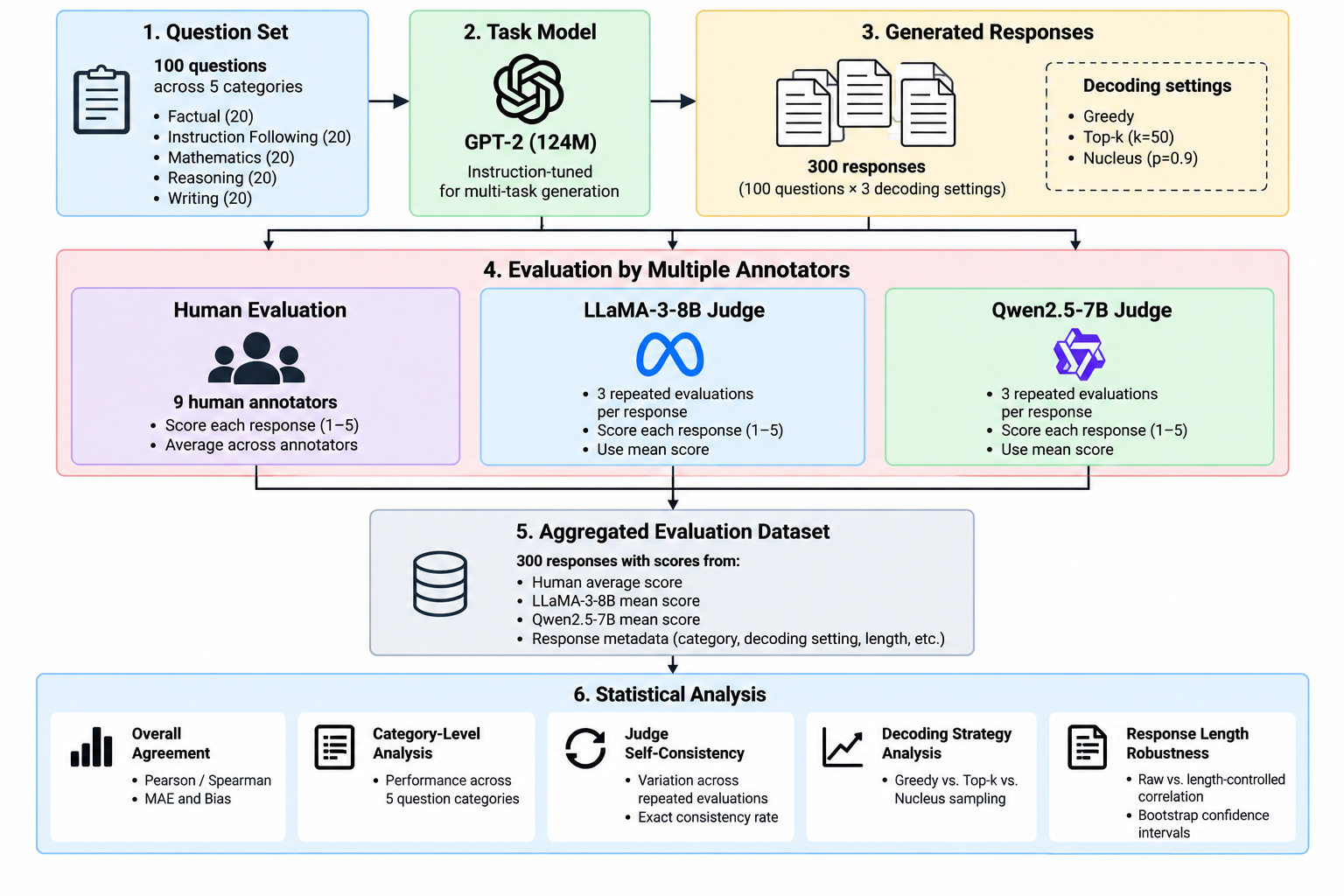}
    \caption{Overall study framework. One hundred questions are used
    to generate 300 responses with an instruction-tuned GPT-2 model.
    The responses are independently evaluated by nine human annotators
    and two local LLM judges using three repeated evaluations. The
    resulting scores are compared using agreement, error, consistency,
    category, decoding, and response-length analyses.}
    \label{fig:study_framework}
\end{figure}

This section describes the experimental design used to evaluate local LLM judges against human ratings, focusing on human alignment and repeated-evaluation consistency. The complete study pipeline is shown in Figure~\ref{fig:study_framework}.

\subsection{Research Questions}
Our study is guided by the following research questions:
\begin{itemize}
    \item \textbf{RQ1:} How strongly do local LLM judges agree with
    independent human ratings of generated responses?
    \item \textbf{RQ2:} How consistent are the judges when the same
    response is evaluated repeatedly using the same protocol?
    \item \textbf{RQ3:} Does human--judge agreement vary across
    question categories and response-generation settings?
    \item \textbf{RQ4:} To what extent does response length explain the
    relationship between human ratings and judge scores?
\end{itemize}

\subsection{Overall Study Design}
The study follows four stages. First, 100 questions covering five evaluation categories are prepared. Second, an instruction-tuned GPT-2 model with 124 million parameters generates three responses per question, producing 300 responses. Third, the responses are evaluated independently by nine human annotators and two local LLM judges, LLaMA-3-8B and Qwen2.5-7B. Finally, human and judge scores are compared
using agreement, error, bias, consistency, category, decoding, and response-length analyses. Each response is evaluated three times by each automated judge, giving
\begin{equation}
300 \times 2 \times 3 = 1800
\end{equation}
automated judge evaluations. Human and automated evaluations are performed independently, and human annotators do not have access to automated judge scores.

\subsection{Dataset and Response Generation}
The evaluation set contains 100 questions divided equally into five categories: factual knowledge, instruction following, mathematics, reasoning, and writing, with 20 questions per category. Three responses are generated for each question using three decoding configurations, producing 300 responses in total and 60 responses per category. The GPT-2 model is instruction-tuned with 124 million parameters. For
each question, it receives the instruction and, when available, an additional input. Generated responses are evaluated without manual correction.

\subsection{Human Evaluation}
Each generated response is independently evaluated by nine human annotators on a 
0--100 scale, where higher scores indicate higher response quality. Annotators evaluate responses with respect to the given instruction without access to decoding configurations or automated judge scores.
For response $i$, let $H_{ij}$ denote the score assigned by human annotator $j$. The human reference score is
\begin{equation}
H_i =
\frac{1}{9}
\sum_{j=1}^{9} H_{ij}.
\end{equation}
Human--human agreement is measured using pairwise Pearson correlation and ICC(2,k), following established principles of annotation reliability \cite{krippendorff2011agreement}.

\subsection{Local LLM Judges and Evaluation Rubric}
Two locally deployed open-weight instruction-tuned models are used as automated judges:
\begin{itemize}
    \item \textbf{LLaMA-3-8B}, deployed locally through Ollama.
    \item \textbf{Qwen2.5-7B}, deployed locally through Ollama.
\end{itemize}
For each response, both judges receive the original instruction, optional input, and generated response, and assign a score from 0 to 100 using the same explicit rubric, referred to as \emph{Judge Prompt v2}. The rubric evaluates:
\begin{enumerate}
    \item correctness,
    \item instruction following,
    \item relevance and completeness,
    \item clarity and coherence, and
    \item writing constraints.
\end{enumerate}
Score anchors are provided for the ranges 0--20, 21--40, 41--60, 61--80, and 81--100. Judges return a final integer score and are not provided with human scores. The same rubric is used for both judges and all repeated evaluations.

\subsection{Repeated Judge Evaluation and Self-Consistency}
Each response is evaluated three times by each automated judge using the same response, instruction, input, and rubric. For judge $m$, response $i$, and repetition $r$, let $J_{im}^{(r)}$ denote the assigned score, where $r \in \{1,2,3\}$.
The final judge score is the mean of the three evaluations:
\begin{equation}
J_{im} =
\frac{1}{3}
\sum_{r=1}^{3} J_{im}^{(r)}.
\end{equation}
Score variability is measured using the standard deviation:
\begin{equation}
SD_{im} =
\sqrt{
\frac{1}{3}
\sum_{r=1}^{3}
\left(J_{im}^{(r)}-J_{im}\right)^2
}.
\end{equation}
Lower standard deviation indicates greater stability. We additionally report exact consistency, defined as the percentage of responses for which all three evaluations produce exactly the same integer score. Self-consistency is evaluated separately from human alignment because stable repeated scores do not necessarily agree with human judgments.

\subsection{Evaluation Metrics}
Human--judge agreement is evaluated using Pearson correlation, Spearman correlation, Mean Absolute Error (MAE), and signed bias. Pearson measures linear association, while Spearman measures rank association between human and judge scores.
MAE measures numerical disagreement:
\begin{equation}
MAE =
\frac{1}{N}
\sum_{i=1}^{N}
|J_i-H_i|.
\end{equation}
Signed bias measures systematic over- or under-scoring:
\begin{equation}
Bias =
\frac{1}{N}
\sum_{i=1}^{N}
(J_i-H_i).
\end{equation}
Lower MAE indicates closer numerical agreement, while positive and negative bias indicate higher and lower judge scores relative to the human reference, respectively. Human--human agreement is reported separately using pairwise Pearson correlation and ICC(2,k).

\subsection{Category and Decoding Analysis}
Pearson correlation and MAE are reported separately for factual knowledge, instruction following, mathematics, reasoning, and writing to examine task-dependent judge behavior. The three GPT-2 decoding configurations are also analyzed separately. Because subgroup sizes are smaller than the complete dataset, these results are interpreted primarily as descriptive evidence.

\subsection{Response Length Analysis}
Response length is measured as the number of characters in each generated response. We calculate Pearson correlations between length and the human reference and between length and each judge score. We then compare the raw human--judge Pearson correlation with a partial Pearson correlation controlling for response length.
For each judge, we report the raw correlation, length-controlled
partial correlation, and their difference. This analysis is treated as a robustness check rather than a causal analysis.

\subsection{Statistical Analysis}
Non-parametric bootstrap resampling is used to quantify uncertainty. For each analysis, 5,000 bootstrap samples are generated by sampling responses with replacement and recomputing the corresponding statistic. We report 95\% bootstrap percentile confidence intervals for the main correlations, MAE, signed bias, and other relevant statistics. The bootstrap random seed is fixed at 20260908 for reproducibility.
For the response-length analysis, bootstrap confidence intervals are also calculated for the difference between the raw and length-controlled human--judge correlations.

\section{Results and Discussion}
This section presents the results for the two local LLM judges, focusing on 
human--judge agreement, self-consistency, category-wise behavior, decoding effects, and response-length robustness. Unless otherwise stated, results are computed over all 300 generated responses.

\subsection{Human--Human Agreement}
Agreement among the nine human annotators is first examined to establish the stability of the human reference. The mean pairwise Pearson correlation is $0.967$, with individual correlations ranging from $0.939$ to $0.990$, and the overall absolute agreement measured using ICC(2,k) is $0.996$. Thus, the human reference is highly consistent, making substantial human--judge disagreement unlikely to be explained primarily by annotator variability.
\begin{table}[H]
\centering
\caption{Agreement among the nine human annotators.}
\label{tab:human_agreement}
\renewcommand{\arraystretch}{1.15}
\begin{tabular}{lc}
\hline
\textbf{Metric} & \textbf{Value} \\
\hline
Number of annotators & 9 \\
Mean pairwise Pearson $r$ & 0.967 \\
Minimum pairwise Pearson $r$ & 0.939 \\
Maximum pairwise Pearson $r$ & 0.990 \\
ICC(2,k) absolute agreement & 0.996 \\
\hline
\end{tabular}
\end{table}

\subsection{Overall Human--Judge Agreement}
\begin{figure}[H]
    \centering
    \includegraphics[width=0.8\textwidth]{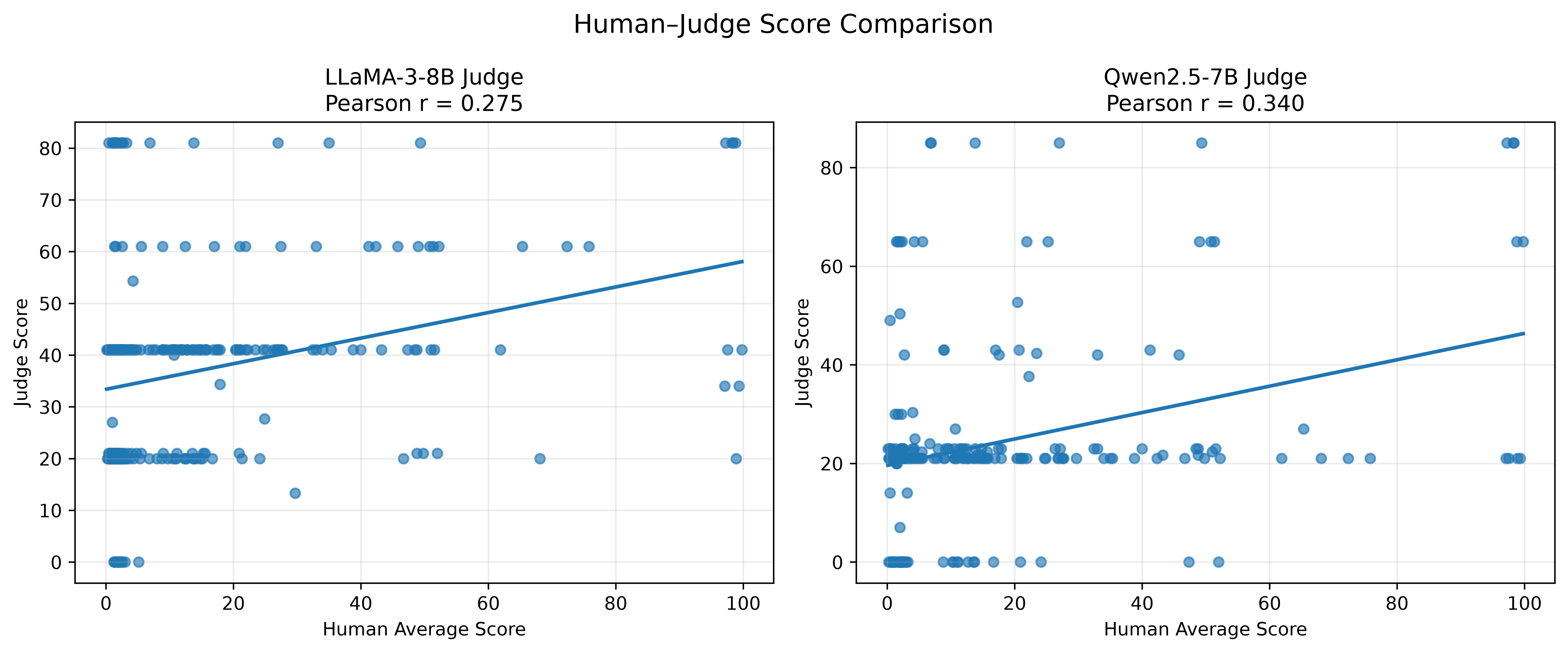}
    \caption{Comparison of human average scores with automated judge
    scores for LLaMA-3-8B and Qwen2.5-7B. Each point represents one
    generated response. The plots illustrate the limited human--judge
    agreement and the positive scoring tendency observed for both
    judges.}
    \label{fig:human_judge_scatter}
\end{figure}

Table~\ref{tab:overall_results} shows positive but limited agreement between both automated judges and the human reference. LLaMA-3-8B achieves Pearson $r=0.275$ and Spearman $r=0.258$, whereas Qwen2.5-7B achieves $0.340$ and $0.303$, respectively. Qwen2.5-7B therefore shows stronger linear and rank-based agreement.
The MAE is $27.71$ for LLaMA-3-8B and $18.64$ for Qwen2.5-7B. Both judges exhibit positive signed bias, $+23.04$ and $+9.54$, respectively, indicating a tendency to assign higher scores than the human reference. The corresponding Pearson 95\% bootstrap confidence intervals are $[0.145,0.405]$ and $[0.178,0.491]$, while the MAE
intervals are $[25.67,29.89]$ and $[16.96,20.45]$.
\begin{table}[H]
\centering
\caption{Overall agreement between human ratings and automated judges.
Confidence intervals are 95\% bootstrap percentile intervals.}
\label{tab:overall_results}
\renewcommand{\arraystretch}{1.15}
\begin{tabular}{lcc}
\hline
\textbf{Metric} & \textbf{LLaMA-3-8B} & \textbf{Qwen2.5-7B} \\
\hline
Pearson correlation & 0.275 & 0.340 \\
Pearson 95\% CI & [0.145, 0.405] & [0.178, 0.491] \\
Spearman correlation & 0.258 & 0.303 \\
Spearman 95\% CI & [0.145, 0.367] & [0.194, 0.405] \\
MAE & 27.71 & 18.64 \\
MAE 95\% CI & [25.67, 29.89] & [16.96, 20.45] \\
Signed bias & +23.04 & +9.54 \\
Bias 95\% CI & [+20.30,+25.72] & [+7.04,+12.00] \\
\hline
\end{tabular}
\end{table}
Figure~\ref{fig:human_judge_scatter} illustrates the limited
human--judge agreement and the positive scoring tendency.
The two automated judges show a Pearson correlation of $0.531$, Spearman correlation of $0.481$, and MAE of $16.86$. Their judge--judge correlation is higher than either judge's correlation with the human reference; however, agreement between automated judges does not establish human-aligned reliability.

\subsection{Self-Consistency of the Judges}
Each response was evaluated three times using the same rubric and
protocol. LLaMA-3-8B has a mean item-level standard deviation of $0.277$ and exact consistency of $97.3\%$, while Qwen2.5-7B has values of $0.395$ and $92.3\%$, respectively. Thus, both judges are highly stable across repeated evaluations. Notably, LLaMA-3-8B has higher exact consistency but lower human--judge correlation and higher MAE, demonstrating that self-consistency and human alignment capture
different properties.
\begin{figure}[H]
    \centering
    \includegraphics[width=0.7\columnwidth]{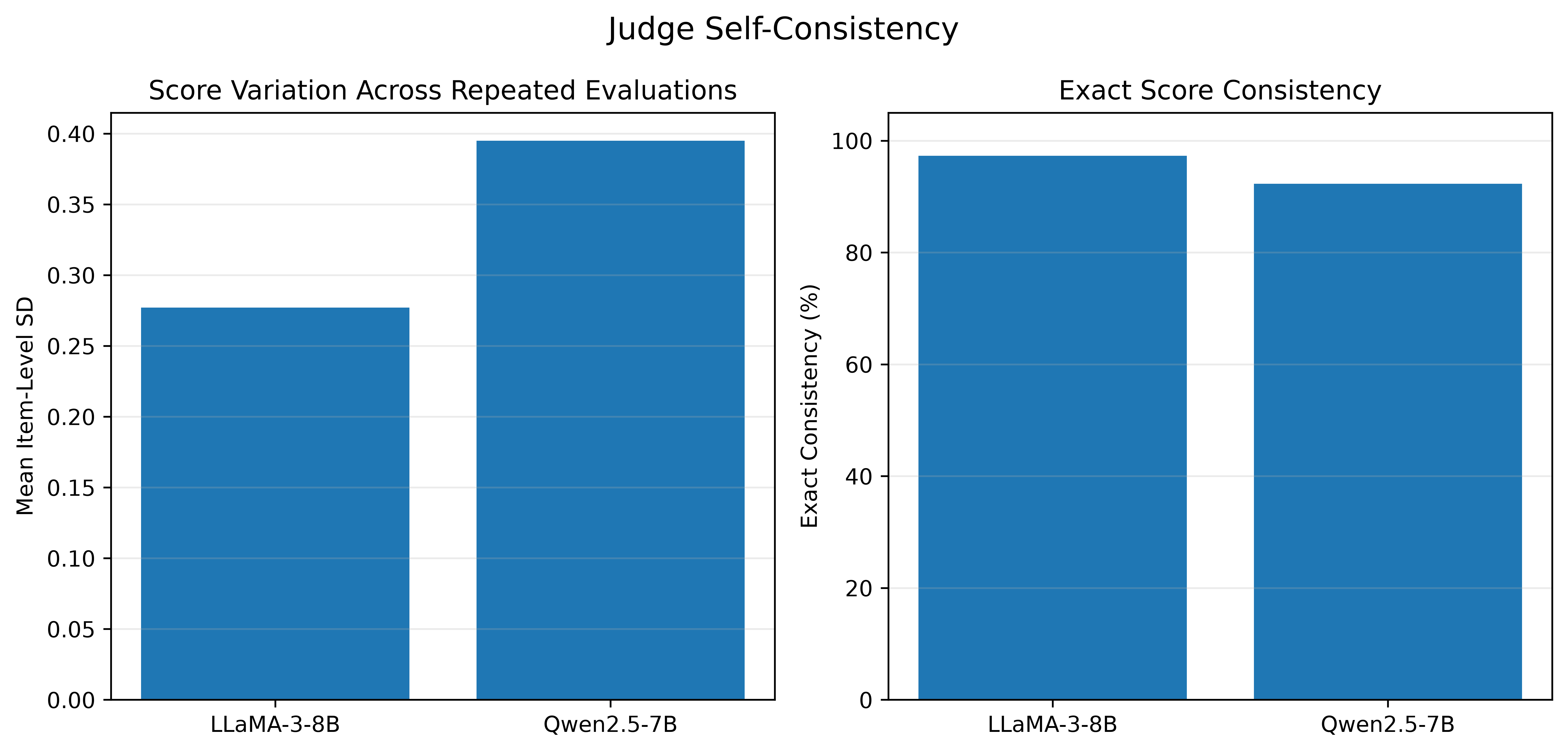}
    \caption{Self-consistency of LLaMA-3-8B and Qwen2.5-7B across
    repeated evaluations. The figure reports mean item-level standard
    deviation and exact consistency.}
    \label{fig:self_consistency}
\end{figure}

\subsection{Category-Wise Performance}
Overall correlations can conceal differences across task types. Table~\ref{tab:category_results} shows that LLaMA-3-8B performs best
on factual knowledge ($r=0.592$) and instruction following ($r=0.462$), but has much lower correlations for mathematics ($0.026$), reasoning ($0.076$), and writing ($0.156$). Qwen2.5-7B also performs best on factual knowledge ($r=0.678$) and shows a relatively strong correlation for mathematics ($0.506$), while its correlations for instruction following, reasoning, and writing are $0.154$, $0.130$, and $0.132$, respectively. The MAE results provide a complementary view. LLaMA-3-8B has its highest MAE for mathematics ($35.95$), whereas Qwen2.5-7B ranges from $16.39$ for instruction following to $21.71$ for writing. These results indicate substantial task dependence in judge behavior.
\begin{table}[H]
\centering
\caption{Human--judge agreement across question categories.}
\label{tab:category_results}
\renewcommand{\arraystretch}{1.15}
\begin{tabular}{lcccc}
\hline
\textbf{Category} &
\multicolumn{2}{c}{\textbf{LLaMA-3-8B}} &
\multicolumn{2}{c}{\textbf{Qwen2.5-7B}} \\
\cline{2-5}
& Pearson $r$ & MAE & Pearson $r$ & MAE \\
\hline
Factual knowledge & 0.592 & 25.79 & 0.678 & 19.46 \\
Instruction following & 0.462 & 24.64 & 0.154 & 16.39 \\
Mathematics & 0.026 & 35.95 & 0.506 & 17.32 \\
Reasoning & 0.076 & 25.52 & 0.130 & 18.34 \\
Writing & 0.156 & 26.64 & 0.132 & 21.71 \\
\hline
\end{tabular}
\end{table}

\subsection{Effect of Response-Generation Decoding}
Human--judge agreement also varies with the decoding configuration used to generate GPT-2 responses. For LLaMA-3-8B, Pearson correlations are $0.264$, $0.372$, and $0.165$ for high-, medium-, and low-temperature settings, with MAEs of $24.33$, $26.81$, and $31.97$, respectively. For Qwen2.5-7B, the corresponding correlations are $0.184$, $0.436$, and $0.314$, with MAEs of $14.74$, $19.13$, and $22.06$,
respectively. The medium-temperature setting produces the highest correlation for both judges. Because each decoding group contains fewer observations than the full dataset, these results are interpreted descriptively.
\begin{table}[H]
\centering
\caption{Human--judge agreement across GPT-2 decoding configurations.}
\label{tab:decoding_results}
\renewcommand{\arraystretch}{1.15}
\begin{tabular}{lcccc}
\hline
\textbf{Setting} &
\multicolumn{2}{c}{\textbf{LLaMA-3-8B}} &
\multicolumn{2}{c}{\textbf{Qwen2.5-7B}} \\
\cline{2-5}
& Pearson $r$ & MAE & Pearson $r$ & MAE \\
\hline
Low temperature & 0.165 & 31.97 & 0.314 & 22.06 \\
Medium temperature & 0.372 & 26.81 & 0.436 & 19.13 \\
High temperature & 0.264 & 24.33 & 0.184 & 14.74 \\
\hline
\end{tabular}
\end{table}

\subsection{Response-Length Analysis}
Across the 300 responses, the mean response length is $129.59$ characters, with a median of $77$ and a range of $3$--$365$ characters. Human scores have a weak positive correlation with response length ($r=0.159$, $p=0.00574$; bootstrap 95\% CI $[0.050,0.290]$). For LLaMA-3-8B, the judge--length correlation is $0.091$
($p=0.1172$; CI $[-0.001,0.182]$). For Qwen2.5-7B, it is $-0.108$ with analytical $p=0.06285$ and bootstrap CI $[-0.195,-0.015]$. Given the difference between the analytical test and bootstrap interval, this is treated as a weak trend rather than
strong evidence of a significant association. Controlling for response length changes the LLaMA-3-8B correlation from $0.275$ to $0.265$, giving $\Delta r=+0.010$ under the raw-minus-partial definition, with bootstrap 95\% CI
$[-0.003,0.031]$. For Qwen2.5-7B, the correlation increases from $0.340$ to $0.364$,
giving $\Delta r=-0.024$ under the same definition, with bootstrap 95\% CI $[-0.051,-0.006]$. Overall, controlling for response length produces only modest changes in the human--judge relationships, indicating that response length
alone does not explain the observed disagreement.

\begin{table}[H]
\centering
\caption{Response-length analysis and length-controlled human--judge
correlations.}
\label{tab:length_analysis}
\renewcommand{\arraystretch}{1.15}
\begin{tabular}{lcc}
\hline
\textbf{Metric} & \textbf{LLaMA-3-8B} & \textbf{Qwen2.5-7B} \\
\hline
Judge--length Pearson $r$ & 0.091 & -0.108 \\
Judge--length $p$-value & 0.1172 & 0.06285 \\
Raw human--judge $r$ & 0.275 & 0.340 \\
Length-controlled $r$ & 0.265 & 0.364 \\
Raw $-$ partial $\Delta r$ & +0.010 & -0.024 \\
\hline
\end{tabular}
\end{table}

\begin{figure}[H]
    \centering
    \includegraphics[width=0.7\columnwidth]{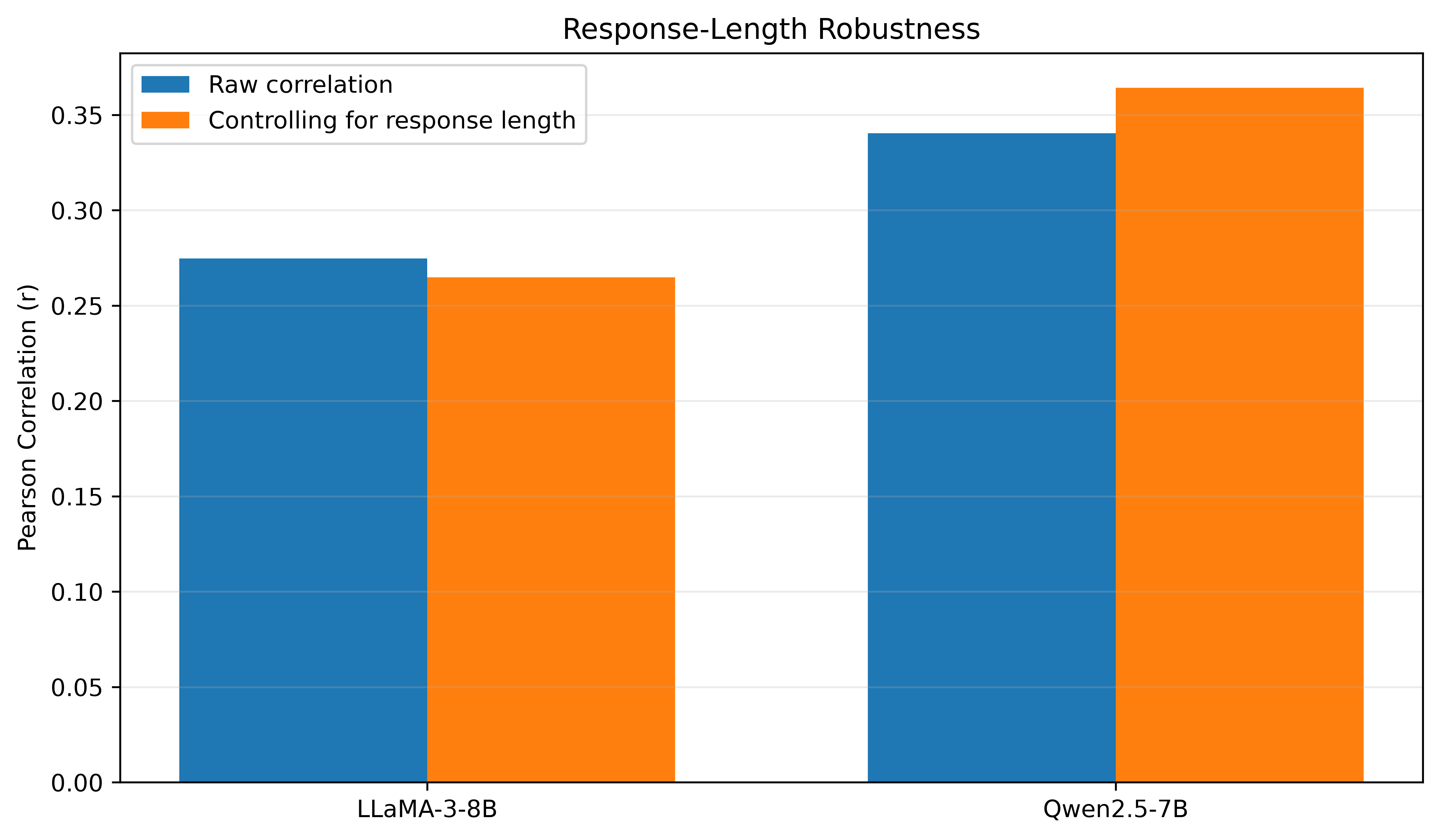}
    \caption{Raw and response-length-controlled Pearson correlations
    between human ratings and automated judge scores. Controlling for
    response length produces only modest changes in the observed
    human--judge relationships.}
    \label{fig:length_robustness}
\end{figure}

\subsection{Discussion of Research Questions}
The results provide answers to all four research questions. Both judges
show positive but limited agreement with human ratings, with Qwen2.5-7B performing better overall than LLaMA-3-8B (Pearson $0.340$ vs.\ $0.275$; MAE $18.64$ vs.\ $27.71$), while both remain highly self-consistent. Judge performance also varies across question categories and decoding configurations. Finally, controlling for response length produces only modest changes in human--judge correlations, indicating that response length alone does not explain the observed disagreement. Overall, the results show that self-consistency and human alignment should be evaluated as separate properties of an automated judge.
\section{Conclusion and Future Work}
This study investigated whether a consistent local LLM judge is necessarily well aligned with human judgments. Using 300 responses generated by an instruction-tuned GPT-2 model, we compared ratings from nine human annotators with those from 
LLaMA-3-8B and Qwen2.5-7B, with each judge evaluating every response three times using the same rubric. The results demonstrate a clear distinction between 
self-consistency and human alignment. Both judges showed high repeatability, but their
agreement with human ratings was substantially lower, and both exhibited positive scoring bias. Judge performance also varied across question categories and decoding configurations, while response-length control produced only modest changes in 
human--judge correlations. Therefore, repeatability alone is not sufficient to establish human-aligned reliability.

\subsection{Future Work}
Future work can extend the evaluation to larger and more diverse datasets, additional local and open-weight judges, and more task categories. Alternative rubrics, pairwise evaluation, and prompting
strategies can be investigated to improve judge calibration and human
alignment. More complex reasoning and domain-specific tasks can also
provide a broader understanding of when local LLM judges can be used
reliably.
\section{Limitations}
This study has several limitations that should be considered when
interpreting the results.
\begin{itemize}
    \item \textbf{Dataset and model scope:} The study uses 100
    questions and 300 responses across five categories, all generated
    by an instruction-tuned GPT-2 (124M) model. Larger and more diverse
    datasets and response-generation models are needed to assess
    broader generalization.
    \item \textbf{Limited judge coverage:} Only LLaMA-3-8B and
    Qwen2.5-7B are evaluated. The observed consistency--alignment
    relationship may differ across other model families and sizes.
    \item \textbf{Evaluation protocol:} Both judges use the same rubric
    and each response is evaluated three times. Alternative rubrics,
    prompts, evaluation formats, or more repetitions may produce
    different results.
    \item \textbf{Controlled setting:} The study focuses on numerical
    scoring against human ratings and does not examine all possible
    LLM-judge biases or real-world evaluation scenarios.
\end{itemize}
These limitations restrict the scope of interpretation and motivate
larger and more diverse evaluations of the consistency--alignment gap.
\section{Ethical Statement and Reproducibility}
\subsection{Ethical Statement}
This study evaluates language model responses using human ratings and
automated LLM-based judges. The evaluation focuses on response quality
and does not require the collection or analysis of sensitive personal
information. Human ratings are used only to construct an aggregate
reference score, and annotator identities are not used in the analysis.

\subsection{Reproducibility}
The study uses a fixed evaluation protocol with the same rubric, repeated evaluation procedure, and 0--100 scoring scale. The evaluation comprises 100 questions, 300 generated responses, nine human annotators, two automated judges, and three repeated evaluations per response. Statistical analysis uses 5,000 bootstrap resamples with a fixed random seed of 20260908. The evaluation procedure, scoring criteria, experimental settings, and analysis methodology are reported to support reproducibility.
\bibliographystyle{tmlr}
\bibliography{references}

\end{document}